\documentclass{article}

\usepackage[nonatbib,final]{nips_2017}

\usepackage[utf8]{inputenc} 
\usepackage[T1]{fontenc}    
\usepackage{hyperref}       
\usepackage{url}            
\usepackage{booktabs}       
\usepackage{amsfonts}       
\usepackage{nicefrac}       
\usepackage{microtype}      
\usepackage{subfigure}
\usepackage{enumitem}

\usepackage[numbers]{natbib}
\usepackage{graphicx}

\title{HoME: a Household Multimodal Environment}

\newcommand*\samethanks[1][\value{footnote}]{\footnotemark[#1]}

\author{
  Simon Brodeur$^1$,
  Ethan Perez$^{2,3}$\thanks{These authors contributed equally.},
  Ankesh Anand$^2$\samethanks,
  Florian Golemo$^{2,4}$\samethanks,\\
  \bf{Luca Celotti$^1$,
  Florian Strub$^{2,5}$,
  Jean Rouat$^1$,
  Hugo Larochelle$^{6,7}$,
  Aaron Courville$^{2,7}$
  }\\
  $^1$Université de Sherbrooke,
  $^2$MILA, Universit\'e de Montr\'eal,
  $^3$Rice University,
  $^4$INRIA Bordeaux,\\
  $^5$Univ. Lille, Inria, UMR 9189 - CRIStAL,
  $^6$Google Brain,
  $^7$CIFAR Fellow
  \\
  \texttt{\{simon.brodeur, luca.celotti, jean.rouat\}@usherbrooke.ca}, \\
  \texttt{ethanperez@rice.edu},
   \texttt{\{florian.golemo, florian.strub\}@inria.fr}, \\
   \texttt{hugolarochelle@google.com},
   \texttt{\{ankesh.anand, aaron.courville\}@umontreal.ca}
}

\begin{document}

\maketitle
\begin{abstract}
We introduce HoME: a \textbf{Ho}usehold \textbf{M}ultimodal \textbf{E}nvironment for artificial agents to learn from vision, audio, semantics, physics, and interaction with objects and other agents, all within a realistic context.
HoME integrates over 45,000 diverse 3D house layouts based on the SUNCG dataset, a scale which may facilitate learning, generalization, and transfer.
HoME is an open-source, OpenAI Gym-compatible platform extensible to tasks in reinforcement learning, language grounding, sound-based navigation, robotics, multi-agent learning, and more.
We hope HoME better enables artificial agents to learn as humans do: in an interactive, multimodal, and richly contextualized setting.
\end{abstract}
\section{Introduction}

Human learning occurs through interaction~\cite{mathematical_kids2012} and multimodal 
experience~\cite{landau1998object,smith2008infants}.
Prior work has argued that machine learning may also benefit from interactive, multimodal learning~\cite{DeepMind-Instruction-Following,Zero-Shot-Deep-Reinforcement-Learning,guesswhat_game}, termed \textit{virtual embodiment}~\cite{Virtual-Embodiment}.
Driven by breakthroughs in static, unimodal tasks such as image classification~\cite{NIPS2012_4824} and language processing~\cite{Word2Vec}, machine learning has moved in this direction.
Recent tasks such as visual question answering~\cite{VQA}, image captioning~\cite{vinyals2017show}, and audio-video classification~\cite{Emotion-Recognition} make steps towards learning from multiple modalities but lack the dynamic, responsive signal from exploratory learning.
Modern, challenging tasks incorporating interaction, such as Atari~\cite{Arcade-Learning-Environment} and Go~\cite{AlphaGo}, push agents to learn complex strategies through trial-and-error but miss information-rich connections across vision, language, sounds, and actions.
To remedy these shortcomings, subsequent work introduces tasks that are both multimodal and interactive, successfully training virtually embodied agents that, for example, ground language in actions and visual percepts in 3D worlds~\cite{DeepMind-Instruction-Following,Zero-Shot-Deep-Reinforcement-Learning,ViZDoom-Instruction-Following}.

For virtual embodiment to reach its full potential, though, agents should be immersed in a rich, lifelike context as humans are.
Agents may then learn to ground concepts not only in various modalities but also in relationships to other concepts, i.e. that forks are often in kitchens, which are near living rooms, which contain sofas, etc.
Humans learn by concept-to-concept association, as shown in child learning psychology~\cite{landau1998object,smith2008infants}, cognitive science~\cite{barsalou2008grounded}, neuroscience~\cite{Hippocampus}, and linguistics~\cite{quine2013word}.
Even in machine learning, contextual information has given rise to effective word representations~\cite{Word2Vec}, improvements in recommendation systems~\cite{adomavicius2011context}, and increased reward quality in robotics~\cite{jaderberg2016reinforcement}.
Importantly, scale in data has proven key in algorithms learning from context~\cite{Word2Vec} and in general~\cite{ILSVRC15,WMT,Simulation-To-Real}.

\begin{figure}[t]
  \centering
  \includegraphics[width=0.95\linewidth]{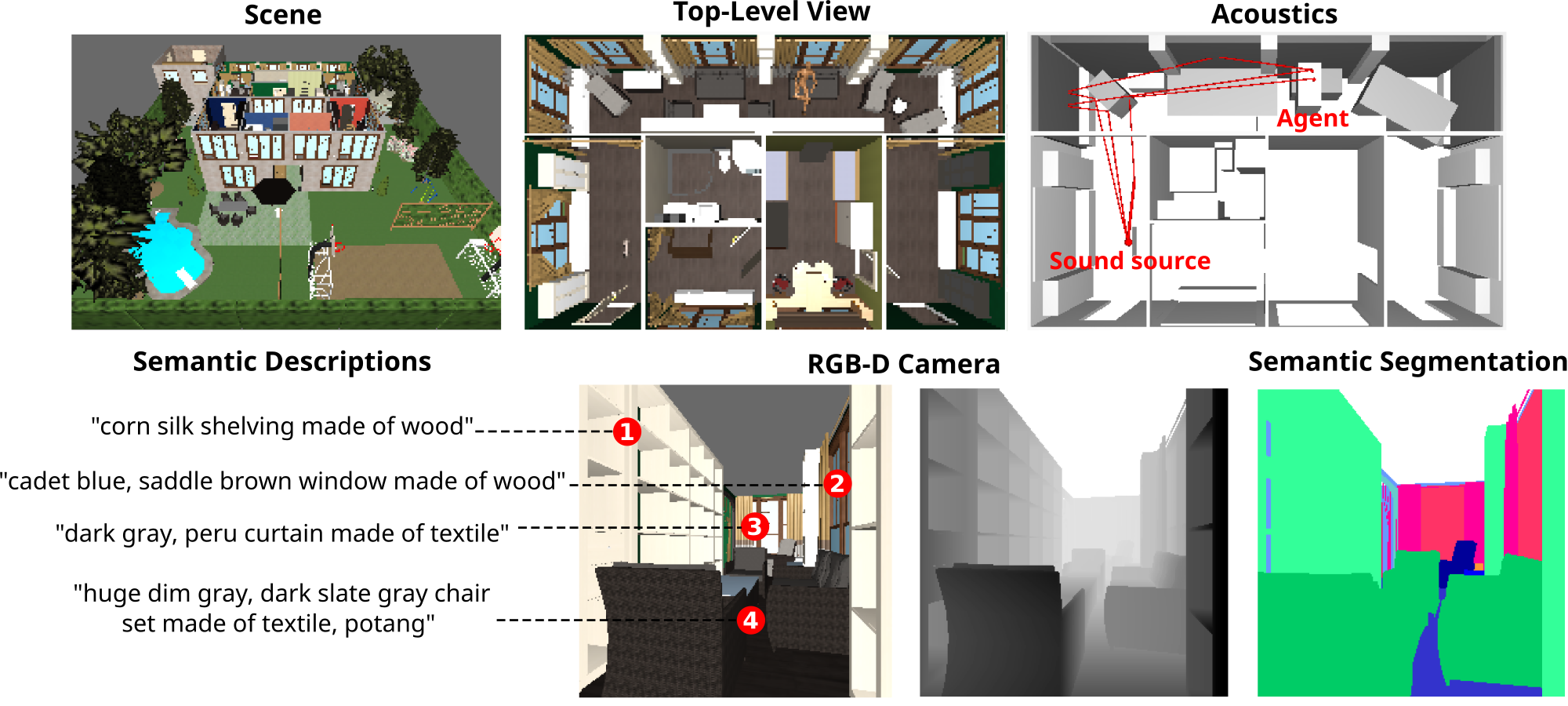}
  \caption{A single example that demonstrates HoME's features.}
  \label{fig:overview}
  \vskip -0.5em
\end{figure}

To this end, we present HoME: the \textbf{Ho}usehold \textbf{M}ultimodal \textbf{E}nvironment (Figure~\ref{fig:overview}). HoME is a large-scale platform\footnote{Available at \url{https://home-platform.github.io/}} for agents to navigate and interact within over 45,000 hand-designed houses from the SUNCG dataset~\cite{SUNCG}. Specifically, HoME provides:
\begin{itemize}[leftmargin=*]
\item 3D visual renderings based on Panda3D.
\item 3D acoustic renderings based on EVERT~\cite{Laine2009}, using ray-tracing for high fidelity audio.
\item Semantic image segmentations and language descriptions of objects.
\item Physics simulation based on Bullet, handling collisions, gravity, agent-object interaction, and more.
\item Multi-agent support.
\item A Python framework integrated with OpenAI Gym~\cite{OpenAI-Gym}.
\end{itemize}
HoME is a general platform extensible to many specific tasks, from reinforcement learning to language grounding to blind navigation, in a real-world context.
HoME is also the first major interactive platform to support high fidelity audio, allowing researchers to better experiment across modalities and develop new tasks.
While HoME is not the first platform to provide realistic context, we show in following sections that HoME provides a more large-scale and multimodal testbed than existing environments, making it more conducive to virtually embodied learning in many scenarios.

\section{Related work}

\begin{figure}[t]
 \centering
  \subfigure{\includegraphics[height=2.5cm]{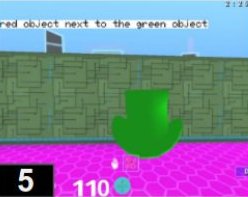}}
  \subfigure{\includegraphics[height=2.5cm]{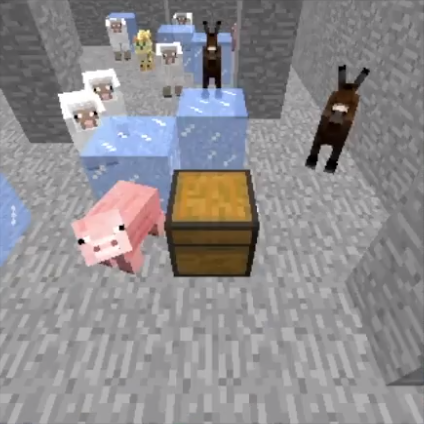}}
  \subfigure{\includegraphics[height=2.5cm]{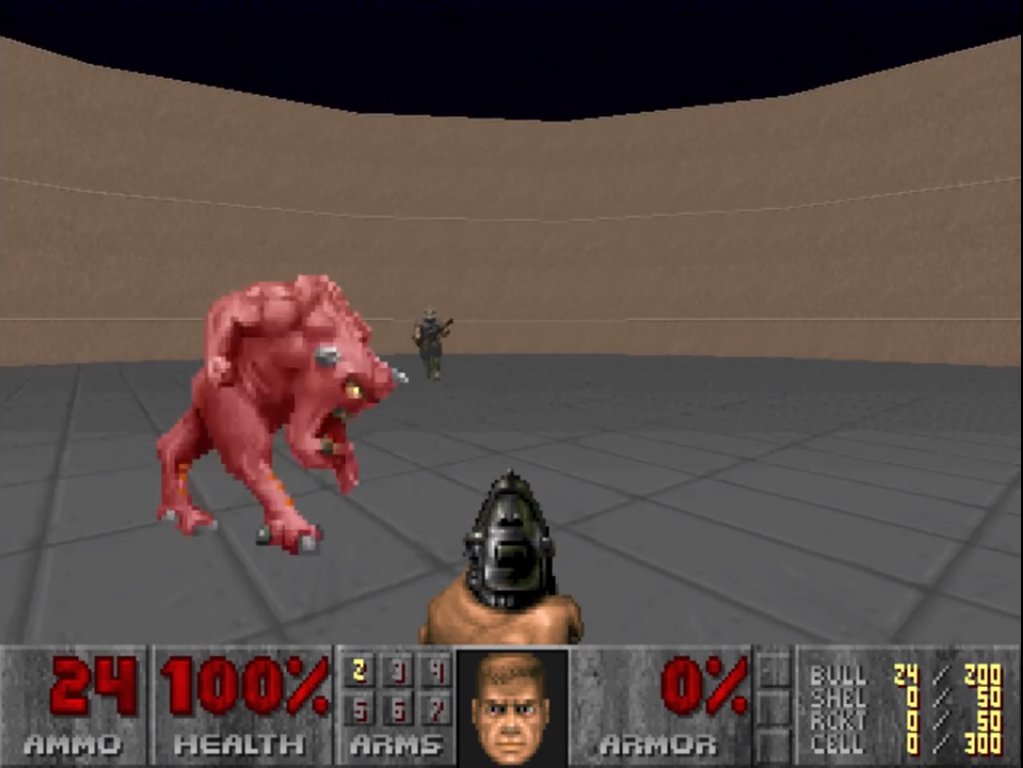}}
  \subfigure{\includegraphics[height=2.5cm]{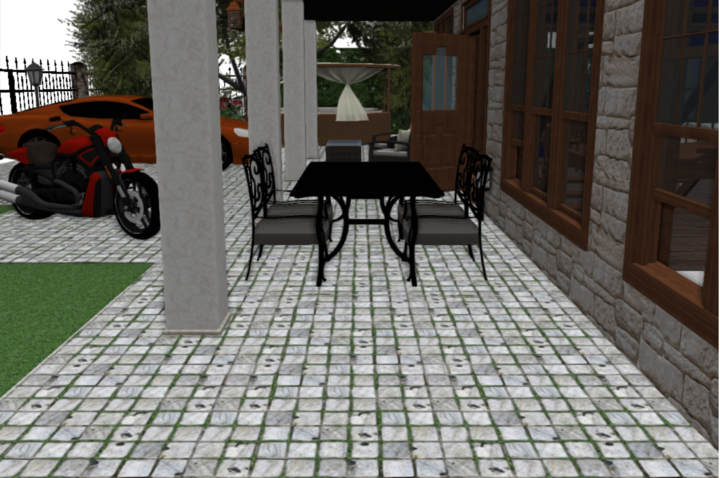}}
  \caption{3D environments (Left to right): DeepMind Lab, Malmo, VizDoom, HoME. Other environments focus on algorithmic challenges; HoME adds a full context in which to learn concepts.}
  \label{fig:comparing-frameworks}
\end{figure}

The AI community has built numerous platforms to drive algorithmic advances: the Arcade Learning Environment~\cite{Arcade-Learning-Environment}, OpenAI Universe~\cite{Universe}, Minecraft-based Malmo~\cite{Malmo}, maze-based DeepMind Lab~\cite{DeepMind-Lab}, Doom-based ViZDoom~\cite{ViZDoom}, AI2-THOR~\cite{AI2-THOR}, Matterport3D Simulator \cite{MatterportSimulator} and House3D~\cite{House3D}.
Several of these environments were created to be powerful 3D sandboxes for developing learning algorithms~\cite{Malmo,DeepMind-Lab,ViZDoom}, while HoME additionally aims to provide a unified platform for multimodal learning in a realistic context (Fig.~\ref{fig:comparing-frameworks}).
Table~\ref{table:comparison} compares these environments to HoME.

The most closely related environments to HoME are House3D, AI2-THOR, and Matterport3D Simulator, three other household environments.
House3D is a concurrently developed environment also based on SUNCG, but House3D lacks sound, true physics simulation, and the capability to interact with objects --- key aspects of multimodal, interactive learning.
AI2-THOR and Matterport3D Simulator are environments focused specifically on visual navigation, using 32 and 90 photorealistic houses, respectively.
HoME instead aims to provide an extensive number of houses (45,622) and easy integration with multiple modalities and new tasks.

Other 3D house datasets could also be turned into interactive platforms, but these datasets are not as large-scale as SUNCG, which consists of 45622 house layouts.
These datasets include Stanford Scenes (1723 layouts)~\cite{fisher2012example}, Matterport3D~\cite{Matterport3D} (90 layouts), sceneNN (100 layouts)~\cite{sceneNN}, SceneNet (57 layouts)~\cite{SceneNet}, and SceneNet RGB-D (57 layouts)~\cite{SceneNet-Trajectories}.
We used SUNCG, as scale and diversity in data have proven critical for machine learning algorithms to generalize~\cite{ILSVRC15,WMT} and transfer, such as from simulation to real~\cite{Simulation-To-Real}.
SUNCG's simpler graphics also allow for faster rendering.

\begin{table}[t]
\centering
\resizebox{\columnwidth}{!}{%
\begin{tabular}{ |l||c|c|c|c|c|c|c|c| }
 \hline
 Environment & 3D & Context & Large-scale & Fast & Customizable & Physics & Acoustics & Photorealistic \\
 \hline\hline 
 Atari~\cite{Arcade-Learning-Environment} &  &  &  & \textbullet &  &  & & \\ \hline
 OpenAI Universe~\cite{Universe} & \textbullet  & \textbullet & \textbullet &  & \textbullet &  & & \\ \hline
 Malmo~\cite{Malmo} & \textbullet &  & \textbullet & \textbullet & \textbullet &  & & \\ \hline
 DeepMind Lab~\cite{DeepMind-Lab} & \textbullet &  &  & \textbullet & \textbullet & &  & \\ \hline
 VizDoom~\cite{ViZDoom} & \textbullet &  &  & \textbullet & \textbullet &  & &\\ \hline
 AI2-THOR~\cite{AI2-THOR} & \textbullet & \textbullet &  & \textbullet &  & \textbullet & & \textbullet\\ \hline
 Matterport3D Simulator~\cite{MatterportSimulator} & \textbullet & \textbullet &  & \textbullet &  & & & \textbullet\\ \hline
 House3D~\cite{House3D} & \textbullet & \textbullet & \textbullet & \textbullet & \textbullet &  & & \\ \hline
\hline
 \textbf{HoME} & \textbullet & \textbullet & \textbullet & \textbullet & \textbullet & \textbullet & \textbullet & \\
 \hline
\end{tabular}
}
\caption{A comparison of modern environments with HoME. \textbf{3D}: Supports 3D settings. \textbf{Context}: Provides a realistic context. \textbf{Large-scale}: Thousands of readily available environments. \textbf{Fast}: Renders quickly. \textbf{Customizable}: Adaptable towards various, specific tasks. \textbf{Physics}: Supports rigid body dynamics and external forces (gravity, collisions, etc.) on agents and objects. \textbf{Acoustics}: Renders audio. \textbf{Photorealistic}: Lifelike visual rendering.}
\label{table:comparison}
\end{table}

\section{HoME}

\begin{figure}[t]
  \centering
  \includegraphics[width=6.2cm]{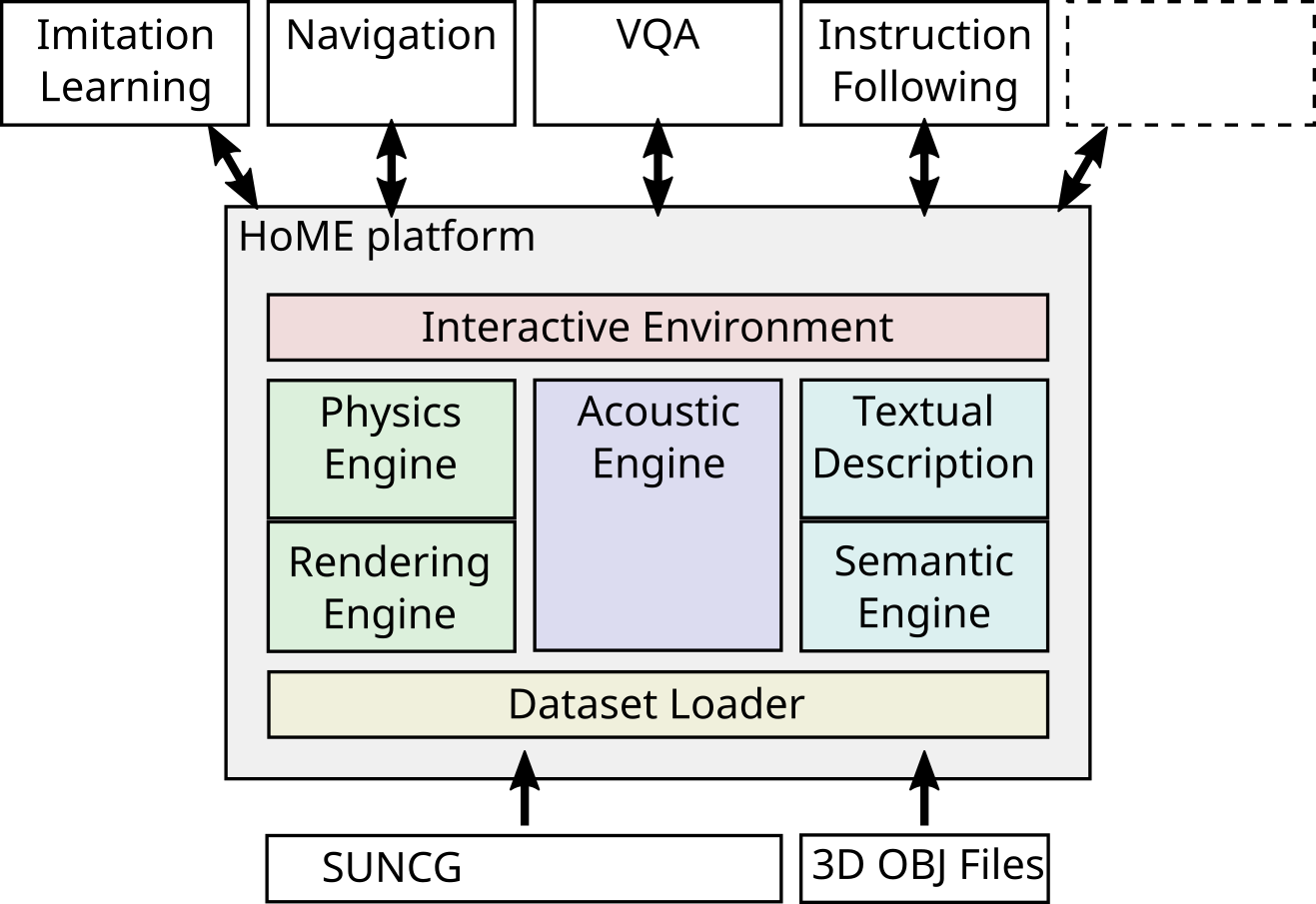}
  \caption{HoME's core components which rely on an underlying dataset loader. These components can be used independently and enable HoME to support various specific tasks.}
  \label{fig:arch-overview}
\end{figure}

Overviewed in Figure~\ref{fig:overview}, HoME is an interactive extension of the SUNCG dataset~\cite{SUNCG}.
SUNCG provides over 45,000 hand-designed house layouts containing over 750,000 hand-designed rooms and sometimes multiple floors.
Within these rooms, of which there are 24 kinds, there are objects from among 84 categories and on average over 14 objects per room.
As shown in Figure~\ref{fig:arch-overview}, HoME consists of several, distinct components built on SUNCG that can be utilized individually.
The platform runs faster than real-time on a single-core CPU, enables GPU acceleration, and allows users to run multiple environment instances in parallel.
These features facilitate faster algorithmic development and learning with more data.
HoME provides an OpenAI Gym-compatible environment which loads agents into randomly selected houses and lets it explore via actions such as moving, looking, and interacting with objects (i.e. pick up, drop, push).
HoME also enables multiple agents to be spawned at once.
The following sections detail HoME's core components.

\subsection{Rendering engine}

The rendering engine is implemented using Panda3D~\cite{Panda3D}, an open-source 3D game engine which ships with complete Python bindings.
For each SUNCG house, HoME renders RGB+Depth scenes based on house and object textures (wooden, metal, rubber, etc.), multi-source lighting, and shadows.
The rendering engine enables tasks such as vision-based navigation, imitation learning, and planning.\\ 
\textbf{This module provides:} RGB image (with different shader presets), depth image.

\subsection{Acoustic engine}

The acoustic engine is implemented using EVERT\footnote{\url{https://github.com/sbrodeur/evert}}, which handles real-time acoustic ray-tracing based on the house and object 3D geometry.
EVERT also supports multiple microphones and sound sources, distance-dependent sound attenuation, frequency-dependent material absorption and reflection (walls muffle sounds, metallic surfaces reflect acoustics, etc.), and air-absorption based on atmospheric conditions (temperature, pressure, humidity, etc.).
Sounds may be instantiated artificially or based on the environment (i.e. a TV with static noise or an agent's surface-dependent footsteps).\\
\textbf{This module provides:} stereo sound frames for agents w.r.t. environmental sound sources.

\subsection{Semantic engine}

HoME provides a short text description for each object, as well as the following semantic information:
\begin{itemize}[leftmargin=*]
\item \textbf{Color}, calculated from object textures and discretized into 16 basic colors, \textasciitilde130 intermediate colors, and \textasciitilde950 detailed colors\footnote{Colors based on a large-scale survey by Randall Munroe, including relevant shades such as ``macaroni and cheese'' and ``ugly pink,'' \url{https://blog.xkcd.com/2010/05/03/color-survey-results/}}.
\item \textbf{Category}, extracted from SUNCG object metadata. HoME provides both generic object categories (i.e. ``air conditioner,'' ``mirror,'' or ``window'') as well as more detailed categories (i.e. ``accordion,'' ``mortar and pestle,'' or ``xbox'').
\item \textbf{Material}, calculated to be the texture, out of 20 possible categories (``wood,'' ``textile,'' etc.), covering the largest object surface area.
\item \textbf{Size} (``small,'' ``medium,'' or ``large'') calculated by comparing an object's mesh volume to a histogram of other objects of the same category.
\item \textbf{Location}, based on ground-truth object coordinates from SUNCG.
\end{itemize}
With these semantics, HoME may be extended to generate language instructions, scene descriptions, or questions, as in~\cite{DeepMind-Instruction-Following,Zero-Shot-Deep-Reinforcement-Learning,ViZDoom-Instruction-Following}.
HoME can also provide agents dense, ground-truth, semantically-annotated images based on SUNCG's 187 fine-grained categories (e.g. bathtub, wall, armchair).\\
\textbf{This module provides:} image segmentations, object semantic attributes and text descriptions.

\subsection{Physics engine}

The physics engine is implemented using the Bullet 3 engine\footnote{\url{https://github.com/bulletphysics/bullet3}}.
For objects, HoME provides two rigid body representations: (a) fast minimal bounding box approximation and (b) exact mesh-based body.
Objects are subject to external forces such as gravity, based on volume-based weight approximations.
The physics engine also allows agents to interact with objects via picking, dropping, pushing, etc.
These features are useful for applications in robotics and language grounding, for instance.\\
\textbf{This module provides:} agent and object positions, velocities, physical interaction, collision.

\section{Applications}
\label{sec:appli}

Using these engines and/or external data collection, HoME can facilitate tasks such as:

\begin{itemize}[leftmargin=*]
\item Instruction Following: An agent is given a description of how to achieve a reward (i.e. ``Go to the kitchen.'' or ``Find the red sofa.'').
\item Visual Question Answering: An agent must answer an environment-based question which might require exploration (i.e. ``How many rooms have a wooden table?'').
\item Dialogue: An agent converses with an oracle with full scene knowledge to solve a difficult task.
\item Pied Piper: One agent must follow another specific agent, out of several, each making specific sounds. HoME's advanced acoustics allow agents with multichannel microphones to perform sound source localization and disentanglement for such a task.
 \item Multi-agent communication: Multiple agents communicate to solve a task and maximize a shared reward. For example, one agent might know reward locations to which it must guide other agents.

\end{itemize}
\section{Conclusion}

Our \textbf{Ho}usehold \textbf{M}ultimodal \textbf{E}nvironment (HoME) provides a platform for agents to learn within a world of context: hand-designed houses, high fidelity sound, simulated physics, comprehensive semantic information, and object and multi-agent interaction.
In this rich setting, many specific tasks may be designed relevant to robotics, reinforcement learning, language grounding, and audio-based learning. HoME's scale may also facilitate better learning, generalization, and transfer.
We hope the research community uses HoME as a stepping stone towards virtually embodied, general-purpose AI.

\section*{Acknowledgments}

We are grateful for the collaborative research environment provided by MILA.
We also acknowledge the following agencies for research funding and computing support: CIFAR, CHISTERA IGLU and CPER Nord-Pas de Calais/FEDER DATA Advanced data science and technologies 2015-2020, Calcul Qu\'{e}bec, Compute Canada, and Google.
We further thank NVIDIA for donating a DGX-1 and Tesla K40 used in this work.
Lastly, we thank acronymcreator.net for the acronym HoME.

{
\small
\bibliography{references}
}
\end{document}